\documentclass[letterpaper, 10 pt, conference]{ieeeconf}
\IEEEoverridecommandlockouts
\usepackage{cite}
\usepackage{amsmath,amssymb,amsfonts}
\usepackage{algorithmic}
\usepackage{graphicx}
\usepackage{textcomp}
\usepackage{xcolor}
\usepackage{mathtools}
\usepackage{ntheorem}
\usepackage{epstopdf}
\usepackage{caption}
\usepackage{setspace}
\usepackage{float}
\usepackage{subcaption}
\usepackage{dsfont}
\usepackage{cleveref}
\usepackage{tabularx,ragged2e,booktabs,caption}
\usepackage[titlenumbered,ruled]{algorithm2e}
\makeatletter
\renewcommand*\env@matrix[1][*\c@MaxMatrixCols c]{%
  \hskip -\arraycolsep
  \let\@ifnextchar\new@ifnextchar
  \array{#1}}
\makeatother

\usepackage{blindtext}

\makeatletter
\newcommand*{\rom}[1]{\expandafter\@slowromancap\romannumeral #1@}
{}
{}
\newtheorem{proposition}{Proposition}{}
\newtheorem{problem}{Problem}{}
{}
{}
{}
{}
\begin{document}
\title{Privacy-Preserving Federated Learning \\ via System Immersion and Random Matrix Encryption}
\author{Haleh Hayati, Carlos Murguia, Nathan van de Wouw% <-this % stops a space
% <-this % stops a space
%\thanks{This work was partially supported by the Australian Research Council (ARC) under the Discovery Project DP170104099.}
\thanks{Haleh Hayati, Carlos Murguia, and Nathan van de Wouw are with the Department of Mechanical Engineering, Dynamics and Control Group, Eindhoven University of Technology, The Netherlands. Emails: \& h.hayati@tue.nl, \& c.g.murguia@tue.nl, \& n.v.d.wouw@tue.nl.}
}
\maketitle

\begin{abstract}
Federated learning (FL) has emerged as a privacy solution for collaborative distributed learning where clients train AI models directly on their devices instead of sharing their data with a centralized (potentially adversarial) server. Although FL preserves local data privacy to some extent, it has been shown that information about clients’ data can still be inferred from model updates.
In recent years, various privacy-preserving schemes have been developed to address this privacy leakage.
However, they often provide privacy at the expense of model performance or system efficiency, and balancing these tradeoffs is a crucial challenge when implementing FL schemes.
In this manuscript, we propose a Privacy-Preserving Federated Learning (PPFL) framework built on the synergy of matrix encryption and system immersion tools from control theory. The idea is to immerse the learning algorithm — a Stochastic Gradient Decent (SGD) — into a higher-dimensional system (the so-called target system) and design the dynamics of the target system so that: trajectories of the original SGD are immersed/embedded in its trajectories; and it learns on encrypted data (here we use random matrix encryption). Matrix encryption is reformulated at the server as a random change of coordinates that maps original parameters to a higher-dimensional parameter space and enforces that the target SGD converges to an encrypted version of the original SGD optimal solution. The server decrypts the aggregated model using the left inverse of the immersion map. We show that our algorithm provides the same level of accuracy and convergence rate as the standard FL with a negligible computation cost while revealing no information about the clients’ data.
\end{abstract}

%\begin{IEEEkeywords}
%Privacy, Stochastic datasets, Mutual Information, Gaussian mechanism.
%\end{IEEEkeywords}

\section{INTRODUCTION}
%% ML and FL
\indent Machine learning (ML) has been successfully used in a wide variety of applications for multiple fields, and industries \cite{jordan2015machine}. In traditional machine learning, training data is centrally held by the server executing the learning algorithm. Distributed learning systems expand this paradigm by branching the learning process to decentralized nodes that only use locally available data.
%However, some learning scenarios must involve less open trust boundaries, particularly when multiple participants are involved.
However, when multiple participants are involved, the exchange of local data poses a significant privacy risk.\\
\indent Federated learning (FL) \cite{mcmahan2017communication, konevcny2016federated,li2020federated} has been recently introduced as a decentralized learning framework that can scale to thousands of participants and preserves data privacy. Its core idea is to train machine learning models on separate datasets distributed across several devices or parties. FL schemes train local models on local clients' datasets, and then clients exchange their parameters (e.g., model weights or gradients) with a central server to aggregate a global model. Since clients do not share their training data, FL is suitable for sensitive data sharing use cases. This includes health care, the Internet of Things, and other scenarios with high privacy concerns \cite{xu2021federated, lu2019blockchain,yang2019federated}. Although FL can provide some level of privacy for clients' raw data, private information can still be inferred from model updates throughout the training process. It has been shown that local models can be traced back to their sources \cite{shokri2017membership, nasr2018comprehensive}. Common attacks to FL are model inversion attacks and gradient inference attacks as identified in \cite{fredrikson2015model,aono2017privacy}.\\\indent
%The information might be leaked, such as by deriving private information from a trained model, model inversion attack, and gradient inference attack as described in \cite{fredrikson2015model,aono2017privacy}.\\
%% who is the adversary?
%In FL, private information can be inferred by two types of actors: internal actors (participating clients and the central server) and external actors (model consumers and eavesdroppers) \cite{yin2021comprehensive}. Because in a typical FL framework, the participant clients collaboratively train a global model using their local datasets, either the clients or the server has the opportunity to access the intermediate training updates (e.g., weights and gradients) and the final model. As a result, certain clients and an honest-but-curious server might be the potential internal adversaries whose objective is to gain access to private data. Furthermore, another type of adversary might come from external actors. Therefore, considering all types of information leakage is necessary to provide privacy in FL.\\
%% other privacy preserving FL works
In recent years, various privacy-preserving schemes have been implemented to address the privacy leakage in FL \cite{mothukuri2021survey,yin2021comprehensive}. Most of them rely on cryptography-based techniques such as Secure Multiparty Computation (SMC) \cite{bonawitz2017practical, mohassel2017secureml, mugunthan2019smpai, so2020scalable,ma2022privacy} and Homomorphic Encryption (HE) \cite{asad2020fedopt}, and perturbation-based techniques such as Differential Privacy (DP) \cite{mcmahan2017communication, geyer2017differentially, bhowmick2018protection, wei2020federated}. Bonawitz et al. \cite{bonawitz2017practical} uses an SMC-based secure aggregation protocol to protect individual model updates
%, in which privacy is provided in local models aggregation. Hence, the server cannot access any local models but still has access to the exact aggregated results.
by aggregating local clients' updates at a trusted party and sharing the aggregated model with the untrusted server.
Although cryptographic algorithms have the advantage of retaining the original accuracy of FL, the resulting solution comes with high additional communication costs. Differential privacy \cite{dwork2006calibrating} is also commonly used to enforce local and global privacy for machine and federated learning. DP provides strong information-theoretic guarantees, is algorithmically simple, and has a small system overhead. However, there is an inherent tradeoff between DP and the performance of federated learning, both in terms of model accuracy and convergence rate, as introducing noise increases privacy but may compromise accuracy dramatically \cite{ wei2020federated }.
%Then, because of the tradeoff between data privacy and utility, this method inevitably incurs a high computational cost.
Because standard cryptographic techniques have a high computation and communication cost, and differential privacy reduces FL performance, in recent years, hybrid privacy-preserving methods that combine cryptographic tools and DP schemes have been proposed to hold acceptable tradeoffs between data privacy and FL performance \cite{truex2019hybrid, xu2019hybridalpha, choquette2021capc}.\\
%%%%%%%%%% our solution
\indent Although current solutions improve privacy of FL, they often do this at the expense of model performance and system efficiency. Balancing these tradeoffs is a key challenge when implementing private FL systems.
%In PPFL algorithms, beyond providing strict privacy guarantees, it is vital to design approaches that are computationally cheap, communication efficient, and without compromising accuracy excessively.\\
It follows that novel Privacy-Preserving FL schemes must be designed to provide strict privacy guarantees, on the one hand, and, on the other hand, have a fair computational cost and use communications efficiently without compromising accuracy excessively.\\
%%%%%%%%%%%%%%%%%%%%%%%%%%%%%%%%%%%%%%%%%%%%%%%%%%%%
\indent In this paper, we propose a Privacy-Preserving Federated Learning (PPFL) framework built on the synergy of matrix encryption and systems immersion tools \cite{astolfi2003immersion} from control theory. The main idea is to treat the learning algorithm used in standard FL — a Stochastic Gradient Decent (SGD) — as a dynamical system that we seek to immerse into a higher-dimensional system (the so-called target system). The dynamics of the target system must be design so that: 1) trajectories of the standard SGD are immersed/embedded in its trajectories; and 2) it learns on encrypted data. We use random matrix encryption, which is reformulated at the server as a random change of coordinates that maps original parameters to a higher-dimensional parameter space and enforces that the target system converges to an encrypted version of the standard SGD optimal solution. The server decrypts the aggregated model using the left inverse of the immersion map.

\indent The proposed framework provides the same accuracy and convergence rate as the standard federated learning (i.e., when no encryption or distortion is induced to protect against data inference), reveals no information about the clients' data, is computationally efficient, and does not degrade the learning performance. To the best of our knowledge, this is the first piece of work that provides a high level of privacy for FL without affecting its performance and excessively increasing communication costs. The main contributions of the paper are summarized as follows: i) using systems immersion tools and random matrix encryption, we develop a privacy-preserving FL scheme that guarantees privacy for local and global models; ii) the proposed scheme is shown to be unconditionally secure \cite{wang2008book}; 
and iii) we validate the performance of the scheme through extensive computer simulations based on a real-world large-scale dataset for a FL network with one server, ten clients, and 199,210 parameters.

%%%%%%%%%%%%%%%%%%%%%%%%%%%%%%%%%%%%%%%%%%%%
\section{Problem Formulation}
\subsection{Standard Federated Learning}
%%%%%%%%%%%%%%%%%%%%%%%%%%%%%%%%%%%%%%%%%%%%%%%%%%%%%%%%%%%%%%%%%%%%%%%%
To develop the architecture of our scheme, we build upon the standard FL algorithm. In standard FL, multiple distributed devices (the clients) and a centralized server aim to train a global AI model without exchanging local data available at the clients. Clients share local model parameters with the server obtained by training a model on their devices using local data. The server aggregates all local models to create a global model that is shared back with clients. Clients use the new global parameters as initial conditions to retrain local models. This procedure is repeated until convergence is achieved \cite{konevcny2016federated}.

Consider a standard FL system with one server and $N_c$ clients. Let $\mathcal{D}_{i}$ denotes the local database held by the $i$-th client, $i \in\{1,2, \ldots, N_c\} =: \mathcal{N}$. At each iteration $t \in \mathbb{N}$, the server broadcasts the latest global model, $w^t \in \mathbb{R}^{n}$ (a vector of parameters), to all clients (starting from a random initial value $w^0$). Iteration times $t \in \mathbb{N}$ are referred to as global iterations. Then, clients determine local AI models, $w_i^t \in \mathbb{R}^{n}$, at their devices by minimizing a given loss function $l(w_i^t,\mathcal{D}_i)$ over local data $\mathcal{D}_{i}$ and the latest update on $w^t$. The latter can be formulated as follows:
\begin{equation}
    w_i^t=\arg \min _{{w}_{i}^{t}} l\left({w}_{i}^{t},\mathcal{D}_i\right). \label{lossfunc}
\end{equation}
Clients send their local optimal $w_i^t$ back to the server and the server updates the global model as follows:
\begin{equation}
  {w}^t=\sum_{i=1}^{N_c} \frac{\left|\mathcal{D}_{i}\right|}{|\mathcal{D}|} {w}^t_{i},\label{serveraggregate}
\end{equation}
where $|\mathcal{D}_i|$ is the size of the $i$-th dataset, $|\mathcal{D}| := \sum_i |\mathcal{D}_i|$, and ${w^t}$ is the global aggregated model. The process is repeated until convergence to the global optimum:
\begin{equation}\label{global}
    {w}^{*}=\arg \min _{w} \sum_{i=1}^{N_c}\frac{\left|\mathcal{D}_{i}\right|}{|\mathcal{D}|} l\left(w,\mathcal{D}_{i}\right).
\end{equation}

In general, standard FL clients use SGD as the optimization algorithm to minimize their local loss function \eqref{lossfunc}. Each client calculates the stochastic gradient of the local model using a mini-batch $\mathcal{X}_{i} \subseteq \mathcal{D}_{i}$ randomly sampled from $\mathcal{D}_{i}$ and updates its local model following $K$ iterations of the SGD:
\begin{equation}
\text{(SGD)} \left\{
\begin{aligned}
&w_{i,0}=w^{t}, \\[1mm]
&w_{i,{k+1}} = {w}_{i,k} -\eta \nabla {l}({w}_{i,k},\mathcal{X}_{i}),  \label{SGDstep}\\[1mm]
&k=0,1, \cdots, K-1,\\[1mm]
&w_{i}^{t+1} = w_{i,K},
\end{aligned}\right.
\end{equation}
where $w_{i,{k}} \in \mathbb{R}^{n}$ denotes the $k$-th local iteration of the SGD algorithm at client $i$, $K$ is the total number of local iterations, and $\eta>0$ is the learning rate of the algorithm. Therefore, at every round, each client initializes the local SGD using the latest received $w^t$ and updates $w_{i}^{t+1}$ via $K$ iterations of the SGD, i.e., $w_{i}^{t+1} = w_{i,K}$. Optimal local parameters, $w_{i}^{t+1}$, are sent to the server for aggregation and the process repeats until convergence. After a sufficient number of global iterations between clients and the server (in the global counter $t$) and local updates (in the local counter $k$), the standard FL scheme converges to the optimal global model \eqref{global} (see \cite{mcmahan2017communication} for details).
%%%%%%%%%%%%%%%%%%%%%%%%%%%%%%%%%%5%%%%%%%%%%%%%%%%%
\subsection{Privacy Requirements}
As discussed in Section $1$, information about participants' private data can still be inferred from the model updates throughout the training process \cite{shokri2017membership, nasr2018comprehensive,fredrikson2015model,aono2017privacy}.
%\cite{shokri2017membership, nasr2018comprehensive}, such as by model inversion attack, and gradient inference attack as described in \cite{fredrikson2015model,aono2017privacy}.
In addition, privacy leakage can also occur in the broadcasting step by analyzing the global model parameters \cite{shokri2017membership}. In FL, there are two types of actors that can infer private information: internal actors (participating clients, the central server, and third parties) and external actors (model consumers and eavesdroppers) \cite{yin2021comprehensive}. We assume all the internal actors are untrusted (honest-but-curious), which means that they will faithfully follow the designed FL protocol but attempt to infer private information. External actors are also untrusted; they aim to eavesdrop the communication between internal actors to infer information. We mainly concentrate on privacy of intermediate local and global models. Privacy of the final model, which will be shared with consumers, can be provided by perturbation-based methods.
%%%%%%%%%%%%%%%%%%%%%%%%%%%%%%%%%%%%%%%%%%%%%%%%%%%%%%%%%%
\subsection{Immersion Map and Target System}
The goal of our privacy-preserving FL scheme is to make inference of the clients' datasets, from the local updates $w_i^t$ and global models $w^t$, as hard as possible without distorting the accuracy and convergence of the learning algorithm (SGD). We aim to design an encryption system through matrix multiplication and system immersion tools from control theory.\\
\indent System immersion refers to embedding the trajectories of a dynamical system into the trajectories of a different higher-dimensional system (the so-called target system) \cite{astolfi2003immersion}. That is, there is a bijection between trajectories of both systems (here referred to as the immersion map), and thus having a trajectory of the target system uniquely determines a trajectory of the original system via the immersion map.\\
\indent In our setting, the idea is to immerse the dynamics of the standard SGD in \eqref{SGDstep} into a target dynamical system -- referred hereafter as the target SGD. The dynamics of the target SGD must be designed so that: 1) trajectories of the standard SGD \eqref{SGDstep} are immersed in its trajectories via a known immersion map; and 2) the target SGD learns on encrypted data (here, we use random matrix encryption). Once we have designed the target system and the immersion map, we can use them to encrypt model updates and learn on encrypted data.\\
%%%%%%%%%%%%%%???????????????\indent We first need to design a random matrix and use it as an encryption key to encrypt aggregated model updates transmitted from the server to the clients. The encrypted global model with encryption key $A^t$ at iteration $t$ is calculated as follows:
%%%%%%%%%%%%%%%%%???????????????????\begin{equation}
%    \tilde{w}^t=A^t w^t, \label{encryption1}
%\end{equation}
%for the random change of coordinates that maps original model parameters to the higher-dimensional parameter space.
%Note that the mapping from the original model parameter vector to the encrypted parameter vector is immersion mapping.
%%%%%%%%%%%%%%where $w^t$ is immersed in $\tilde{w}^t$. Then,
\indent Consider the original vector of parameters $w_{i,k} \in \mathbb{R}^n$ in \eqref{SGDstep} at time $k$, and denote the vector generated by the target SGD as $\tilde{w}_{i,k} \in \mathbb{R}^m$ with $m>n$. Consider the following general target SGD:
\begin{equation}
\text{(Target SGD)} \left\{
\begin{aligned}
&\tilde{w}_{i,0}=\tilde{w}^{t} \\[1mm]
&\tilde{w}_{i,{k+1}}= f(\tilde{w}_{i,k}),\label{targetSGDstep}\\[1mm]
&k=0,1, \cdots, K-1,\\[1mm]
&\tilde{w}_{i}^{t+1} = \tilde{w}_{i,K},
\end{aligned}\right.
\end{equation}
with function $f:\mathbb{R}^m \rightarrow \mathbb{R}^m$ to be designed and initial condition $\tilde{w}^{t}$ (the latest encrypted global update from the server). We say that the standard SGD in \eqref{SGDstep} is immersed in the target SGD \eqref{targetSGDstep}, if there exists a left invertible function $\pi:\mathbb{R}^n \to \mathbb{R}^m$ satisfying:
\begin{equation}
\tilde{w}_{i,k}=\pi\left(w_{i,k}\right), \label{immersionmapping}
\end{equation}
for all $k \in \mathcal{K}:=\{1,2, \cdots, K-1\}$. We refer to this function $\pi(\cdot)$ as the \emph{immersion map}. Because \eqref{immersionmapping} must be satisfied for all $k \in \mathcal{K}$, we need to enforce (by designing $\pi(\cdot)$ and $f(\cdot)$) that: \textbf{(a)} the initial condition of \eqref{targetSGDstep}, $\tilde{w}_{i,0}=\tilde{w}^{t}$, satisfies $\tilde{w}_{i,0} = \pi(w_{i,0}) = \pi(w^t)$, where $w^t$ is the latest \emph{unencrypted} global update shared by the server (i.e., the update that would be produced by the standard FL scheme in \eqref{serveraggregate}); and \textbf{(b)} the dynamics of both algorithms match under the immersion map, i.e., $\tilde{w}_{i,k+1}=\pi\left(w_{i,k+1}\right)$. Condition (a) implies that what the server sends to clients is $\pi(w^t)$. That is, \emph{the immersion map $\pi(\cdot)$ is the encryption scheme for all clients.} It follows that the first constraint on $\pi(\cdot)$ is that it must comply with the privacy requirements. Next, using the expressions for $w_{i,k+1}$ and $\tilde{w}_{i,k+1}$, in \eqref{SGDstep} and \eqref{targetSGDstep}, respectively, and \eqref{immersionmapping}, condition (b), $\tilde{w}_{i,k+1}=\pi\left(w_{i,k+1}\right)$, can be written as follows:
\begin{equation}
f(\pi\left(w_{i,k}\right)) = \pi\left({w}_{i,k} -\eta \nabla {l}({w}_{i,k},\mathcal{X}_{i})\right), \label{immersioncondition}
\end{equation}
which is a time-varying nonlinear equation on ${w}_{i,k}$. We refer to this equation as the \emph{immersion condition}.
\subsection{Secure Aggregation and Problem Statement}
Once a complete cycle has been finished by the target SGD, so $k = K-1$, all clients send their last iteration, $\tilde{w}_{i,K}$, to a third party for data aggregation. We refer to this party simply as the \emph{aggregator}. The role of the aggregator is to interface between clients and the server and thus prevent the server from accessing exact local models. The aggregator takes the updated encrypted local models from all clients, $\tilde{w}_{i,K}$, aggregates them and sends the aggregated model to the server. Consequently, the server cannot access any local model and only has access to the aggregated results. Moreover, since the aggregator has access to the encrypted local updates $\tilde{w}_{i,K}$ only, it is not required to be trusted. The aggregated model at the $t$-th iteration is given by
\begin{equation}
   \tilde{w}^{t+1}=\sum_{i=1}^{N_c} \frac{|\mathcal{D}_i|}{|\mathcal{D}|} \tilde{w}_{i,K} = \sum_{i=1}^{N_c} \frac{|\mathcal{D}_i|}{|\mathcal{D}|} \pi\left(w_{i,K}\right) \label{distortedaggregatedModel},
\end{equation}
where the right-hand side part of \eqref{distortedaggregatedModel} follows from the immersion condition (b).

The server receives $\tilde{w}^{t+1}$ in \eqref{distortedaggregatedModel} and aims to retrieve $w^{t+1} = \sum_{i=1}^{N_c} (|\mathcal{D}_i|/|\mathcal{D}|) w_{i,K}$ -- the aggregated result of the standard SGD in \eqref{SGDstep}. The latter imposes an extra condition on the immersion map, $\pi(\cdot)$, since to retrieve $w^{t+1}$ from $\tilde{w}^{t+1}$: \textbf{(c)} there must exists a function $\pi^L:\mathbb{R}^m \to \mathbb{R}^n$ satisfying the following left-invertibility condition:
\begin{equation} \label{left_inverse}
\pi^L \left( \sum_{i=1}^{N_c} \frac{|\mathcal{D}_i|}{|\mathcal{D}|} \pi\left(w_{i,K}\right) \right) = \sum_{i=1}^{N_c} \frac{|\mathcal{D}_i|}{|\mathcal{D}|} w_{i,K}.
\end{equation}
If such $\pi^L(\cdot)$ and $\pi(\cdot)$ exist, the server can retrieve the original aggregated parameters $\sum_{i=1}^{N_c} (|\mathcal{D}_i|/|\mathcal{D}|) w_{i,K}$ by passing the encrypted aggregated results through function $\pi^L(\cdot)$. We have now all the machinery required to state the problem we seek to solve.

\begin{problem}\emph{\textbf{(Privacy-Preserving FL)}} Consider the standard SGD \eqref{SGDstep} and the target SGD \eqref{targetSGDstep}. Design an immersion map $\pi(\cdot)$ and function $f(\cdot)$ in \eqref{targetSGDstep} so that: \textbf{(a)} the initial condition of \eqref{targetSGDstep}, $\tilde{w}_{i,0}=\tilde{w}^{t}$, satisfies $\tilde{w}_{i,0} = \pi(w_{i,0}) = \pi(w^t)$; \textbf{(b)} the dynamics of both algorithms match under the immersion map, i.e., the immersion condition \eqref{immersioncondition} is satisfied; and \textbf{(c)} there exists a function $\pi^L(\cdot)$ satisfying \eqref{left_inverse}.
\end{problem}
%%%%%%%%%%%%%%%%
\begin{figure*}
  \centering
  \includegraphics[width=0.85\textwidth]{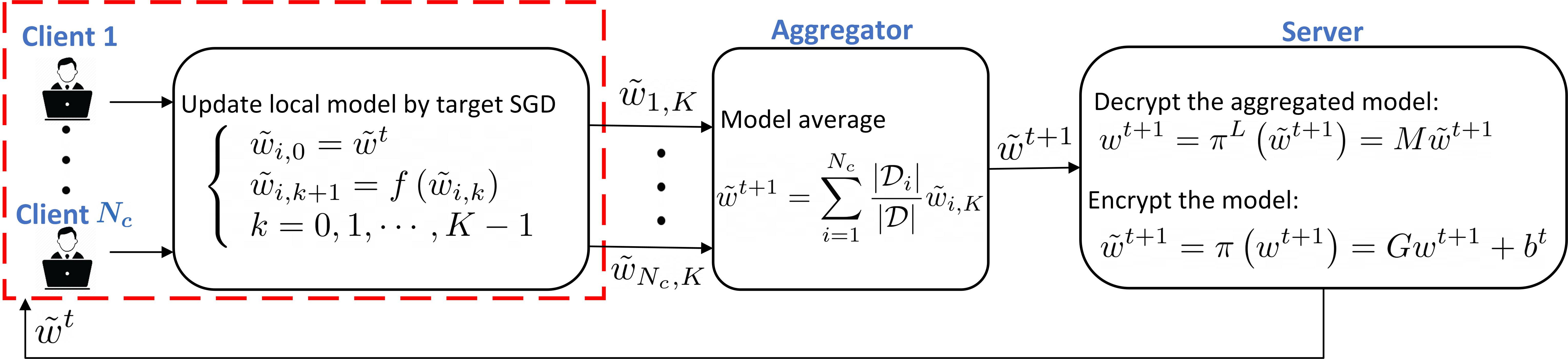}
    \caption{Flowchart of SIFL.}
    \label{flowchart}
\end{figure*}
\section{Solution to Problem 1}
\subsection{Solution}
In this section, we introduce the proposed privacy-preserving FL algorithm. We construct this algorithm by deriving particular solutions to all functions in Problem 1.\linebreak Since the problem formulation and solution are based on systems immersion theory, we refer to our algorithm as \emph{System Immersion based Federated Learning} (SIFL). We start with function $f(\cdot)$ and condition \textbf{(b)}, i.e., the immersion condition \eqref{immersioncondition}. A natural candidate for $f(\tilde{w}_{i,k})$ of the target SGD dynamics \eqref{targetSGDstep} is a gradient-dependent function. Let $f(\tilde{w}_{i,k})$ have the following form
\begin{equation}
\tilde{w}_{i,{k+1}}= f(\tilde{w}_{i,k}) := \tilde{w}_{i,k}-\eta \widetilde{\nabla l}(\tilde{w}_{i,k},\mathcal{X}_{i}),\label{eq1}
\end{equation}
where $\eta>0$ and $\mathcal{X}_{i}$ are the same learning rate and data realization as in the standard SGD \eqref{SGDstep}, and $\widetilde{\nabla l}(\tilde{w}_{i,k},\mathcal{X}_{i})$ is a gradient function to be designed. With this $f(\tilde{w}_{i,k})$, the immersion condition \eqref{immersioncondition} takes the form
\begin{equation}
    \begin{aligned}
    &\pi\left(w_{i,k}\right) - \eta \widetilde{\nabla l}(\pi\left(w_{i,k}\right),\mathcal{X}_{i})\\ &\hspace{30mm}= \pi\left({w}_{i,k} -\eta \nabla {l}({w}_{i,k},\mathcal{X}_{i})\right).
    \end{aligned}\label{eq2}
\end{equation}
Let the immersion map $\pi(\cdot)$ be an affine function (as privacy mechanism in \cite{hayati2021finite})
\begin{equation}
\pi(s) := Gs+b^t, \label{mapping}
\end{equation}
for some matrix $G \in \mathbb{R}^{m \times n}$ and $b^t \in \mathbb{R}^{m}$ -- with slight abuse of notation, we let $b^t$ change with the global counter $t$ independently of the argument $s$. Then, the immersion condition reduces to
\begin{equation}
 \widetilde{\nabla l}\left(G w_{i,k}+b^t,\mathcal{X}_{i}\right)=G \nabla l\left(w_{i,k},\mathcal{X}_{i}\right).\label{eq4}
\end{equation}
Finally, let the modified gradient function be of the form $\widetilde{\nabla l}(\tilde{w}_{i,k},\mathcal{X}_{i})=G \nabla l(M \tilde{w}_{i,k},\mathcal{X}_{i})$, for some $M \in \mathbb{R}^{n \times m}$ to be designed. Hence, the immersion condition takes the form
\begin{equation}
G \nabla l\left(M\left(G w_{i,k}+b^t\right),\mathcal{X}_{i}\right)=G \nabla l\left(w_{i,k},\mathcal{X}_{i}\right). \label{eqc1}
\end{equation}
To satisfy \eqref{eqc1}, we must have $M\left(G w_{i,k}+b^t\right)=w_{i,k}$, i.e., $MG=I$ and $b^t \in \text{ker}[M]$. It follows that: 1) $G$ must be of full column rank ($\text{rank}[G]=n$); 2) $M$ is a left inverse of $G$ (which always exists given the rank of $G$); and 3) $b^t \in \text{ker}[M]$ and this kernel is always nonempty because $M$ is full row rank by construction. So the final form for $f(\cdot)$ in \eqref{targetSGDstep} is given as
\begin{equation}
f(\tilde{w}_{i,k}) = \tilde{w}_{i,k} - \eta G{\nabla l}(M\tilde{w}_{i,k},\mathcal{X}_{i}),\label{eqc2}
\end{equation}
with $\nabla l(\cdot)$ and $\eta$ the gradient and learning rate of the standard SGD in \eqref{SGDstep}, and $(G,M)$ as defined above.\\
\indent Note that substitution of the immersion map \eqref{mapping} into \eqref{immersionmapping} leads the solution of the target system \eqref{eqc2} to be an affine function of trajectories of the original SGD as follows:
\begin{equation}
   \tilde{w}_{i,K}=\pi(w_{i,K})=G w_{i,K} + b^t. \label{distortedUpdatedModel}
\end{equation}
%where $\tilde{w}_{i,{t+1}}$ and $w_{i,{t+1}}$ denote the updated local models of $i$-th client by target SGD and original SGD, respectively.\\
By plugging in the designed immersion map \eqref{mapping} in the aggregated encrypted model \eqref{distortedaggregatedModel} yields 
\begin{align}
     \tilde{w}^{t+1}&= \sum_{i=1}^{N_c} \frac{|\mathcal{D}_i|}{|\mathcal{D}|} \left(G w_{i,K} + b^t\right) \nonumber\\
     &= G \left(\sum_{i=1}^{N_c} \frac{|\mathcal{D}_i|}{|\mathcal{D}|} w_{i,K}\right) + b^t\nonumber\\
     &=G w^{t+1} + b^t,\label{distortedaggregatedModel2}
\end{align}
where $\tilde{w}^{t+1}$ and $w^{t+1}$ denote the aggregated encrypted and unencrypted updated models, respectively.\\
\indent We have designed the function $f(\cdot)$ and the immersion map $\pi(\cdot)$ to satisfy the immersion condition \eqref{immersioncondition}. Next, we seek for a function $\pi^L(\cdot)$ satisfying \eqref{left_inverse} (condition \textbf{(c)} in Problem 1). Given \eqref{distortedaggregatedModel2}, condition \eqref{left_inverse} can be written as
\begin{align} \label{left_inverse2}
\pi^L \left( G \left(\sum_{i=1}^{N_c} \frac{|\mathcal{D}_i|}{|\mathcal{D}|} w_{i,K}\right) + b^t \right) =  \sum_{i=1}^{N_c} \frac{|\mathcal{D}_i|}{|\mathcal{D}|} w_{i,K},
\end{align}
which trivially leads to
\begin{equation}
    \pi^L(s) := Ms,\label{inversemap}
\end{equation}
since $MG=I$ and $b^t \in \text{ker}[M]$. Finally, condition \textbf{(a)}  in Problem 1 is automatically satisfied for the designed functions as the initial condition of \eqref{targetSGDstep}, $\tilde{w}_{i,0}=\tilde{w}^{t}$, is what the server sends to clients at global iteration $t$, and the server encrypts the aggregated result with the immersion map, i.e, $\tilde{w}_{i,0} = \pi(w^t) = \pi(w_{i,0})$. 

At every global iteration $t$, the server designs a vector $b^t$ satisfying $Mb^t=0$ and uses it to construct the immersion map $\pi(s) = Gs+b^t$ (the encryption scheme). To increase security, the server uses this $b^t$ to add randomness to the mapping by exploiting the nonempty kernel of $M$. We let $b^t$ be of the form $b^t=NR^t$ for some matrix $N \in \mathbb{R}^{m \times (m-n)}$ expanding the kernel of $M$ (i.e., $MN=0$) and some random vector $R^t \in \mathbb{R}^{(m-n) \times 1}$. Hence, we have $Mb^t=0$ in all iterations and $b^t=NR^t$ changes randomly with $t$.
\begin{proposition}[Solution to Problem 1]: The immersion map $\pi(\cdot)$, target SGD function $f(\cdot)$, and function $\pi^L(\cdot)$:
\begin{equation}
    \left\{\begin{array}{l}
\tilde{w}_{i,k}=\pi\left(w_{i,k}\right)=G w_{i,k} + N R^t, \\[2mm]
f(\tilde{w}_{i,k}) = \tilde{w}_{i,k} - \eta G{\nabla l}(M\tilde{w}_{i,k},\mathcal{X}_{i}),\\[2mm]
\pi^L(\tilde{w}^{t+1}) = M\tilde{w}^{t+1}={w}^{t+1},\label{immersionsolution}
\end{array}\right.
\end{equation}
provide a solution to Problem 1.
\end{proposition}
\emph{\textbf{Proof}}: The proof follows from the analysis provided in the solution section above, Section \rom{3}.
\hfill $\blacksquare$
\subsection{SIFL Algorithm}
The flowchart of the SIFL is shown in Figure \ref{flowchart}. The summary of the algorithm is as follows:
\begin{itemize}
\item \textbf{FL initialization and encryption by the server}. The server initializes the global model $w^0$ and encrypts it as $\tilde{w}^0=
G w^0+b^0$. Then, it immerses the original SGD into the target SGD as \eqref{eqc2} and broadcasts $\tilde{w}^0$, target SGD, and other hyperparameters to clients.
\item \textbf{Local model training and update by clients}. The clients receive the current encrypted global model $\tilde{w}^t$ sent by the server and update their individual local model parameters using their local datasets $\mathcal{D}_i$ and the target SGD system \eqref{eqc2}. Then, they send their updated model to the aggregator for aggregation.
\item \textbf{Global model aggregation}. The aggregator takes the average of local encrypted models and sends the aggregated model \eqref{distortedaggregatedModel2} to the server.
\item \textbf{Global model encryption and broadcasting by the server}. The server decrypts the aggregated global model using function $\pi^L(\cdot)$ in \eqref{inversemap}. Then, it encrypts the new global model using the immersion map $\pi(\cdot)$ \eqref{mapping} and broadcasts it to all clients for the next round.
\end{itemize}
The pseudo-code of SIFL is shown in Algorithm $1$.
%\textcolor[rgb]{0.00,0.50,0.00}{The pseudo-code of SIFL is shown in Algorithm $1$. In this privacy-preserving algorithm, the server sends the target SGD and the encrypted global model to clients. Hence, clients don't have access to the original SGD and global models. Clients iterate the encrypted model on immersed SGD (target system) and send the encrypted local model $\tilde{w}^t_i$ to the aggregator. Aggregator sends the aggregated model $\tilde{w}^t$ back to the server. Since $w^t$ is immersed in $\tilde{w}^t$, the server can extract original $w^t$ via the left inverse of the immersion map.}
%\textbf{\textcolor[rgb]{1.00,0.00,0.00}{This sounds redundant again, do you need this paragraph? You just wrote a summary and gave an algorithm.
%}}
%%%%%%%%%%%%%%%%%% Algorithm
\begin{algorithm}
    \caption{System Immersion based FL algorithm}\label{alg:one}
    \SetKwInOut{KwIn}{Input}
    \SetKwInOut{KwOut}{Output}
\KwIn{Set of clients and their databases $\mathcal{D}_i$, number of FL iterations $T$, learning rate $\eta$, number of local SGD iterations $K$, privacy matrix $M_{n \times m}$, its right inverse $G$, and matrix $N$ in its null space.}
%\Output{distorted model $\tilde{w}^t$}
\bigskip
\textbf{Handshaking phase:}\\
The server sends target SGD \eqref{eqc2}, the encrypted initialized global model $\tilde{w}^0$, and other hyperparameters to clients for model update.\\
\bigskip
\textbf{Server Execution:}\\

\For{each FL iteration $t$ from $0$ to $T-1$}{
    \For{$i$-th client}{
        $\tilde{w}_{i,K} \gets$ \textbf{ClientUpdate$\left(\tilde{w}^{t},\mathcal{D}_{i}\right)$}}
The aggregator aggregates local models
\begin{equation}
   \tilde{w}^{t+1}=\sum_{i=1}^{N_c} \frac{|\mathcal{D}_i|}{|\mathcal{D}|} \tilde{w}_{i,K}
   =G w^{t+1} + N R^{t}. \nonumber
\end{equation}
The aggregator sends $\tilde{w}^{t+1}$ to the server\\
     ${w}^{t+1}=\pi^L(\tilde{w}^{t+1})=M \tilde{w}^{t+1}$\tcp*{Server decryption}
     Server creates random vector $R^t$.\\
    $\tilde{w}^{t+1}=\pi(w^{t+1})=Gw^{t+1}+NR^{t+1}$\tcp*{Server encryption}
    Server sends encrypted $\tilde{w}^{t+1}$ to clients.}
\bigskip
\textbf{ClientUpdate$\left(\tilde{w}^{t},\mathcal{D}_{i}\right)$:} \tcp*{Run on client $i$}
$\mathcal{X}_i \leftarrow\left(\right.$ split $\mathcal{D}_{i}$ into batches $\left.\right)$\\
Initialize: $\tilde{w}_{i,0} \gets \tilde{w}^{t}$\\
\For{ each local epoch $k$ from 1 to $(K-1)$}{
\For{ batch $x_i \in \mathcal{X}_i$}{
$\tilde{w}_{i,k+1} \leftarrow \tilde{w}_{i,k} -\eta G \nabla {l}(M \tilde{w}_{i,k},x_i)$\\
%${w}_i^{t+1} \leftarrow {w}^t -\eta \nabla {l}({w}^t , x_i)$
}
}
\Return{$\tilde{w}_{i,K}=G w_{i,K} + NR^{t}$} to the aggregator.
\end{algorithm}
%%%%%%%%%%%%%%%%%%%%%%%%%
%%???????????????\indent In this framework, the server cannot access any local model and only have access to the aggregated models. Clients and the aggregator do not have access to each round's original local and global models. Therefore, all the models transmitting in communication network are encrypted.
%% scheme 2
%%%%%%%%%%
%%%%%%%%%%%%%%%%%%%%%%% algorithm 2
%%%%%%%%%%%%%%%%%%%%%%%%%%%%%%%%%%%%%%%%%%%%%%%%%%%
\section{Security Analysis}
In SIFL, since local and global models are encrypted in all rounds, even if adversaries wish to gain data by attacking the server or a communication channel, they can only get the encrypted models. Even in the case of internal adversaries, they require to break the cryptosystem to access the data. Hence, neither internal adversaries nor external ones (model consumers and eavesdroppers) have access to the original local models. Therefore, they need to break the cryptosystem to infer information about clients’ data. Since the encryption keys are random and changed at each iteration, even if adversaries are lucky enough to break some rounds of training results, they cannot access the actual data due to the inclusion of randomness.\\
\indent In \cite{ shannon1949communication}, Shannon proves that a necessary condition for an encryption method to be unconditionally secure is that the uncertainty of the secret key is larger than or equal to the uncertainty of the plaintext \cite{wang2008book}. He proposes a one-time pad encryption scheme in which the key is randomly selected and never used again. The one-time pad gives unbounded entropy of the key space, i.e., infinite key space, which provides unconditional security \cite{menezes2018handbook,diffie2019new} (the unconditional secrecy would be lost when the key is not random or if it is reused). Since in SIFL the encryption keys are random and only used once, it provides infinite key space, and thus, it can be considered unconditionally secure.
%%%%%%%%%%

\section{SIMULATION RESULTS}
\begin{figure}[!htb]
  \centering
  \includegraphics[width=3.5in]{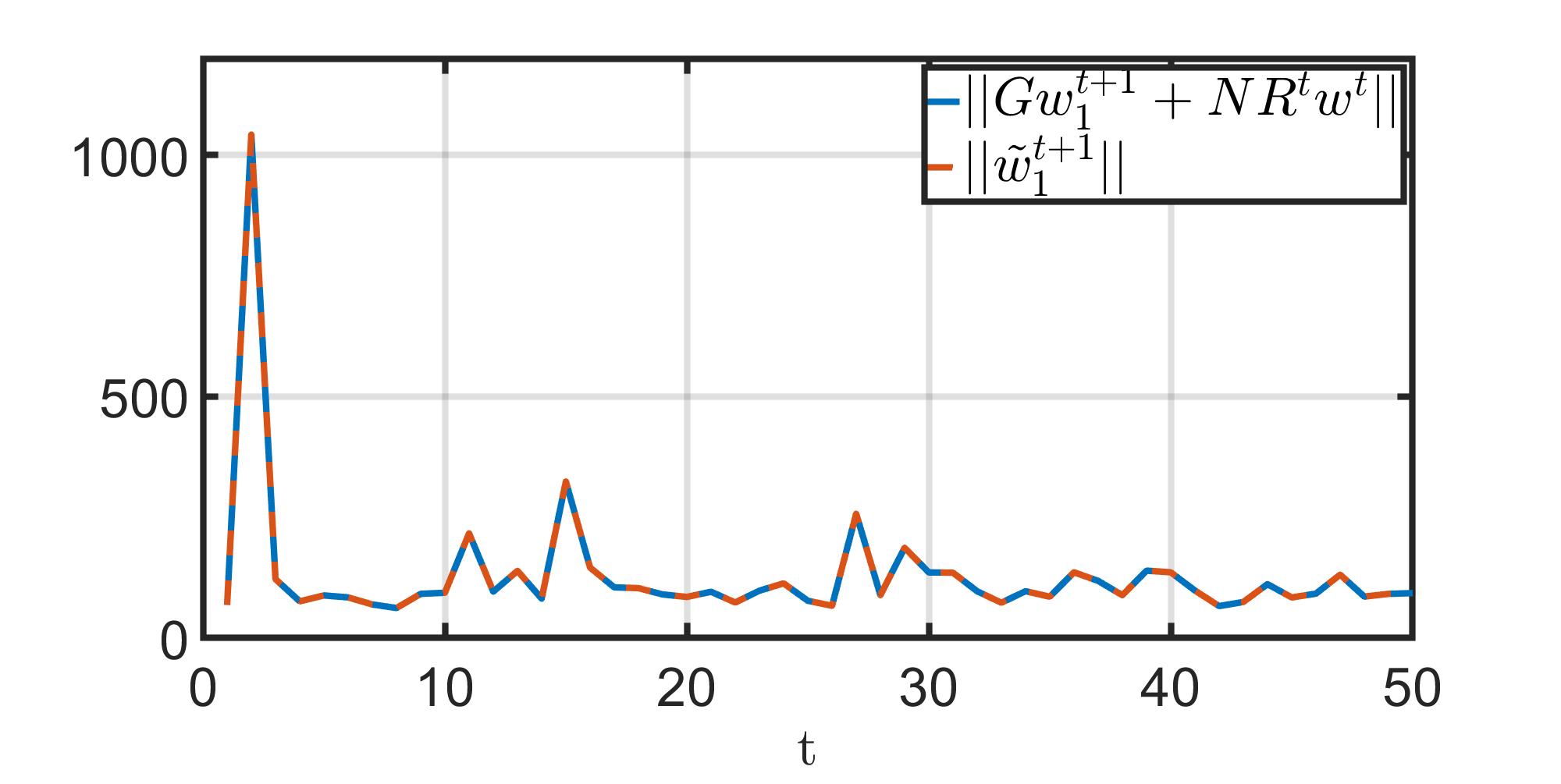}
  \caption{The comparison of the norm of the encrypted local model parameters of the first client, $||\tilde{w}_1^{t+1}||$ and its expected amount $|| G w_1^{t+1} + NR^t||$.}\label{fig:Normpflocal model}
\end{figure}

\begin{figure}[!htb]
  \centering
  \includegraphics[width=3.5in]{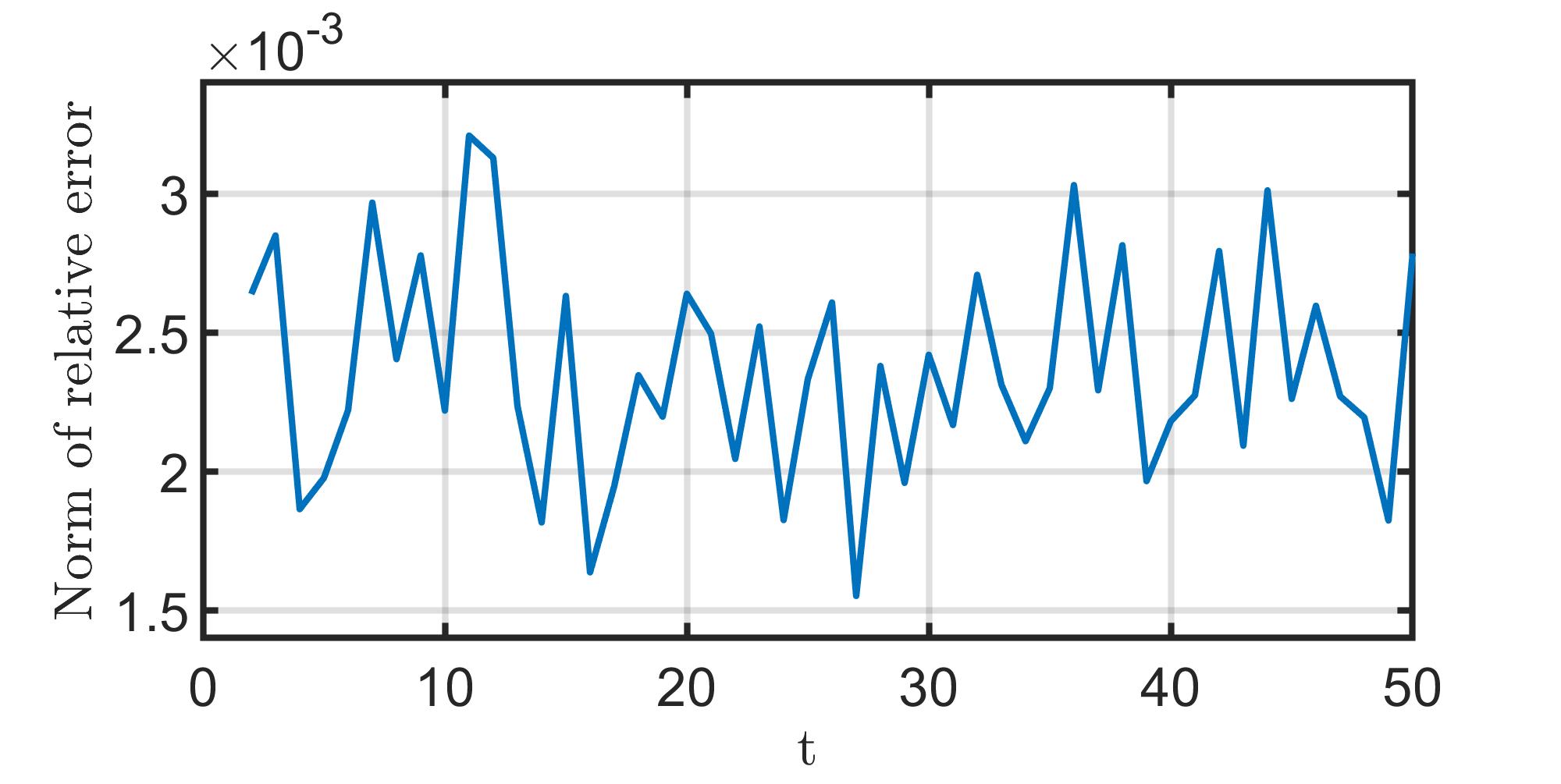}
  \caption{The norm of the relative error between encrypted local model parameters $||\tilde{w}_1^{t+1}||$ and $|| G w_1^{t+1} + NR^t||$.}\label{fig:errorlocalmodel}
\end{figure}

\begin{figure}[!htb]
  \centering
  \includegraphics[width=3.5in]{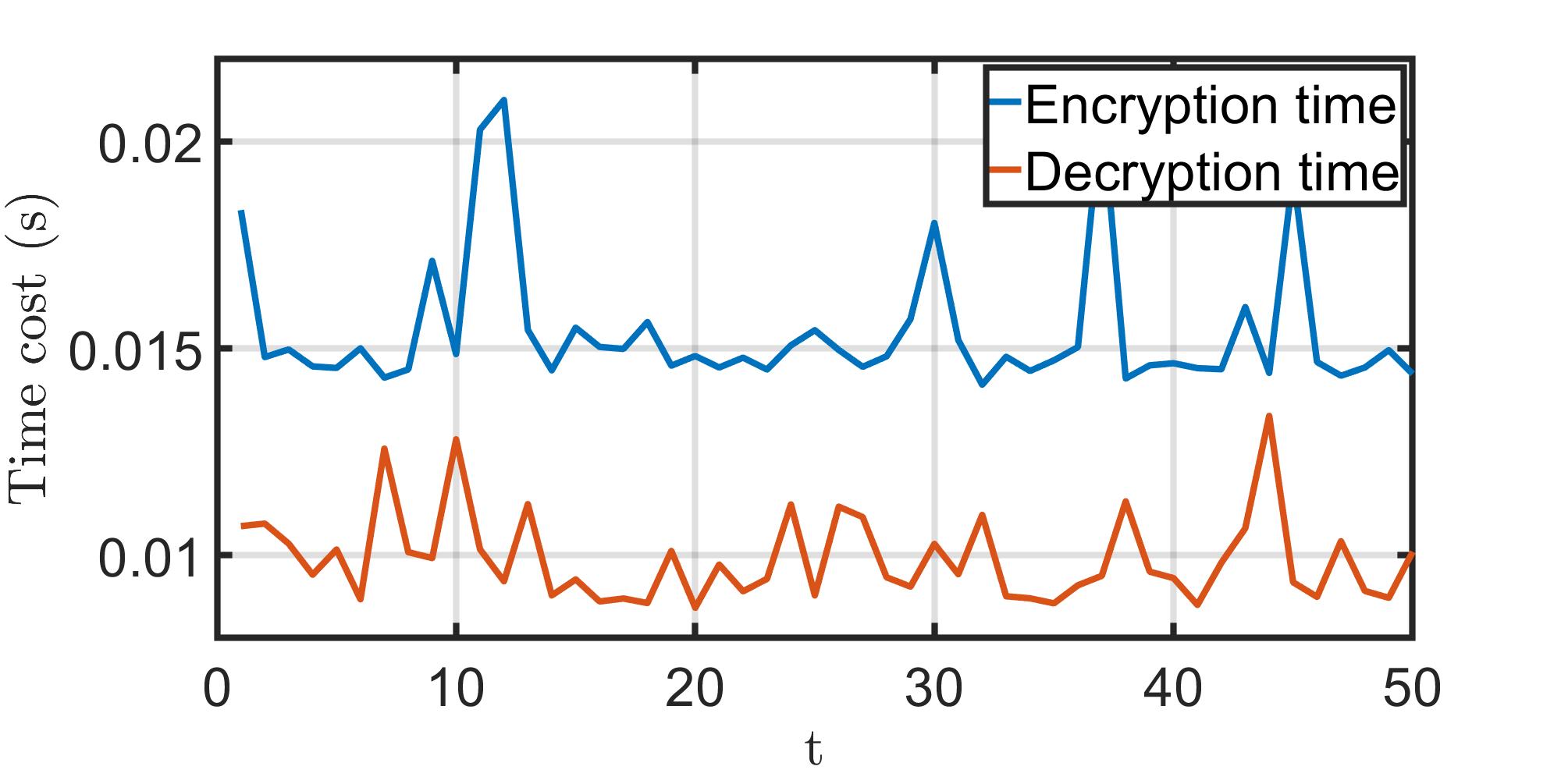}
  \caption{The encryption and decryption time cost in each iteration of SIFL.} \label{fig:encryptdecrypttime}
\end{figure}

\begin{figure}[!htb]
  \centering
  \includegraphics[width=3.5in]{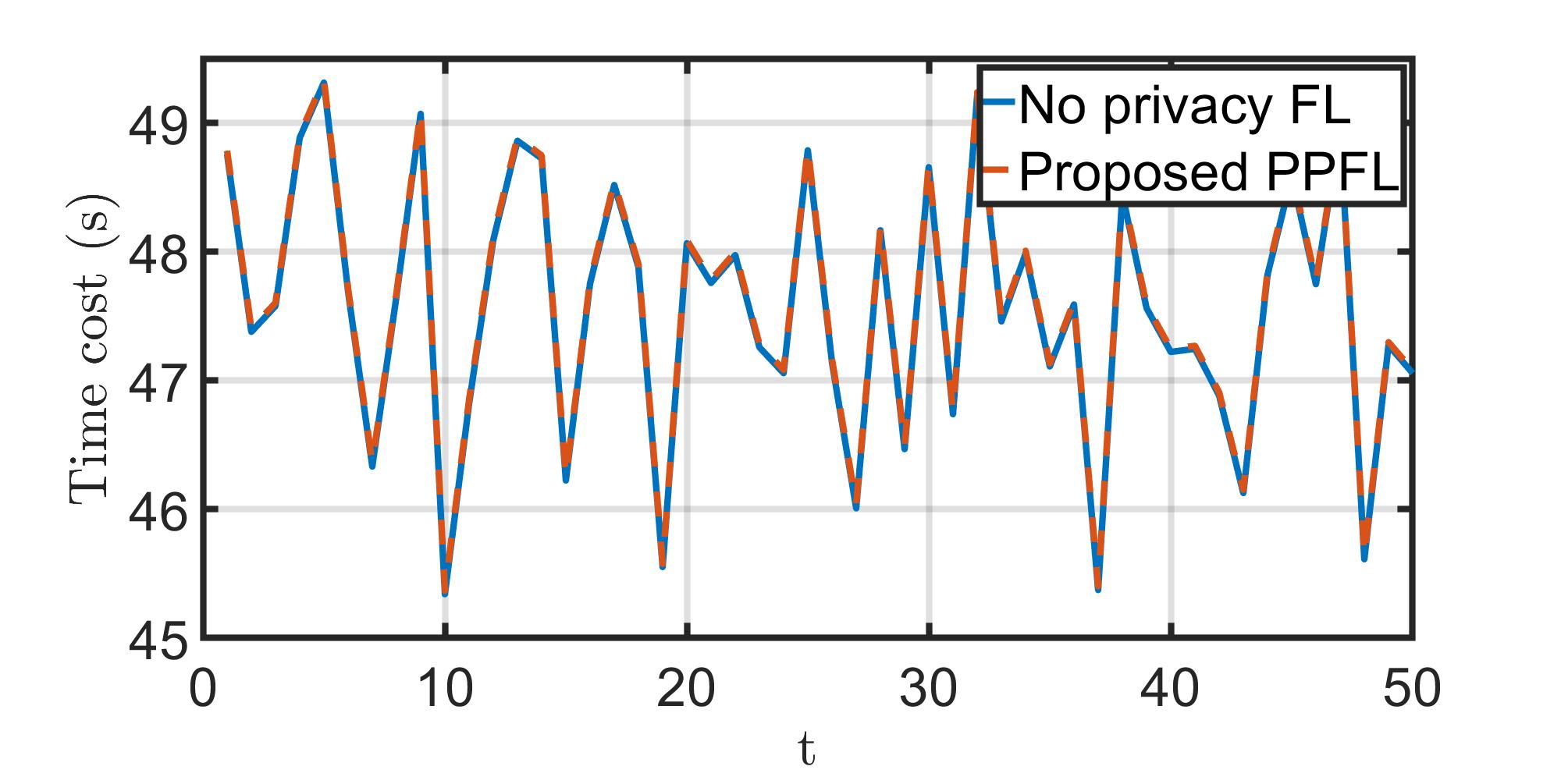}
  \caption{The comparison of the training time of FL with and without the proposed privacy mechanism.}\label{fig:timecost}
\end{figure}

In this section, we implement our proposed scheme for performance evaluation using multi-layer perception (MLP) \cite{tang2015extreme} and a real-world federated dataset. Since FL is mostly suited for parameterized learning, such as all types of neural networks, MLP is employed for the learning method. We test our algorithm on the standard MNIST dataset for handwritten digit recognition, containing 60000 training and 10000 testing instances of 28× 28 size gray-level images  \cite{lecun1998gradient}. Our model uses an MLP network with two hidden layers containing 200 hidden units. This feed-forward neural network uses ReLU units and softmax of 10 classes (corresponding to the ten digits) and with ten clients. For the network optimizer, we consider cross-entropy loss and SGD optimizer with the learning rate 0.01 and the local epoch $K = 2$. To assess model quality, we used the pre-defined MNIST test set. Our implementation uses Keras with a Tensorflow backend.\\
%%%%%%%%%%%%%%%%%%%%%%%%%%
\begin{figure}[!htb]
  \centering
  \includegraphics[width=3.5in]{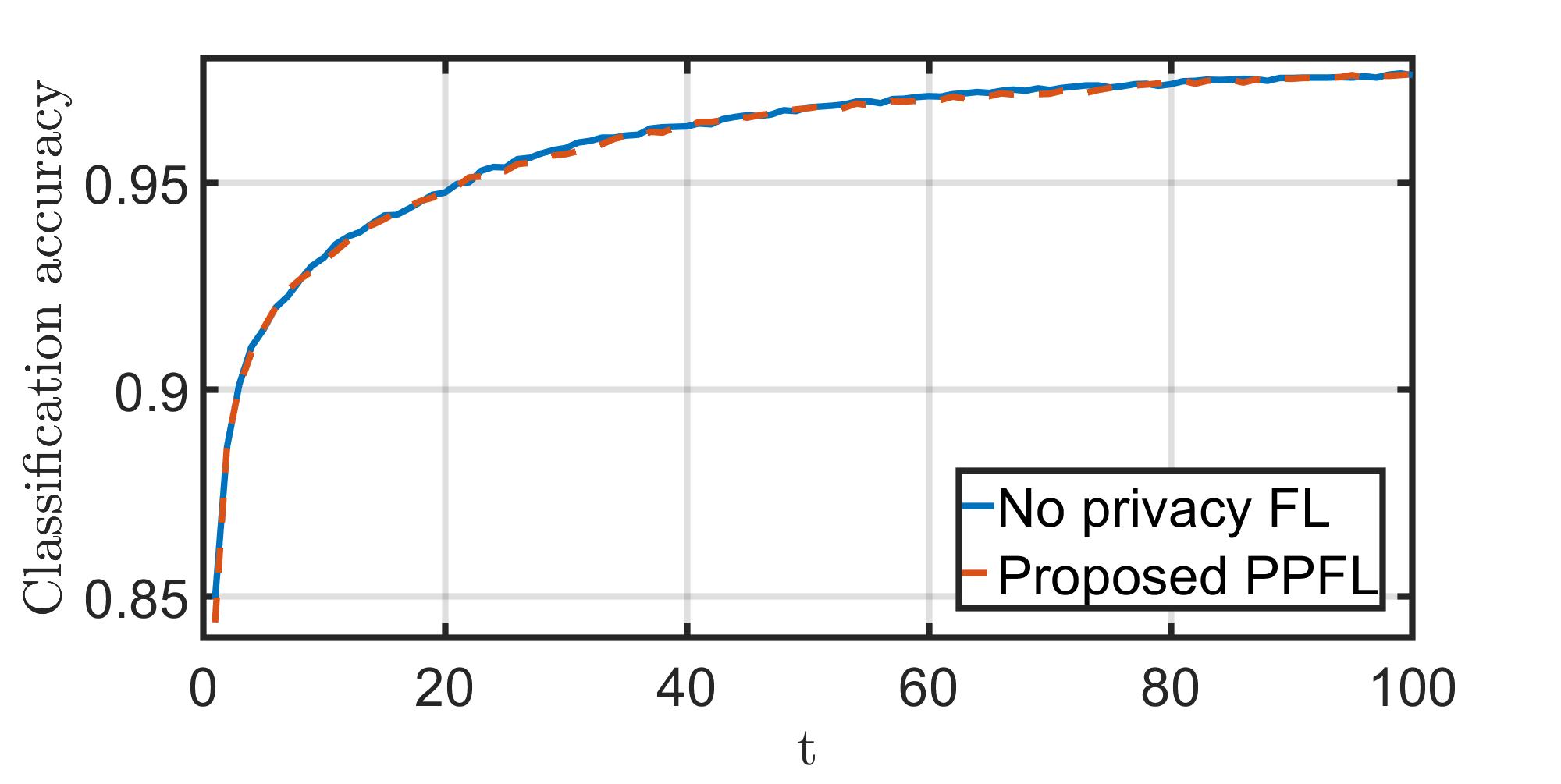}
  \caption{The comparison of the accuracy of FL network in each global iteration $t$ with and without the proposed privacy mechanism.}\label{fig:accuracy}
\end{figure}
\begin{figure}[!htb]
  \centering
  \includegraphics[width=3.5in]{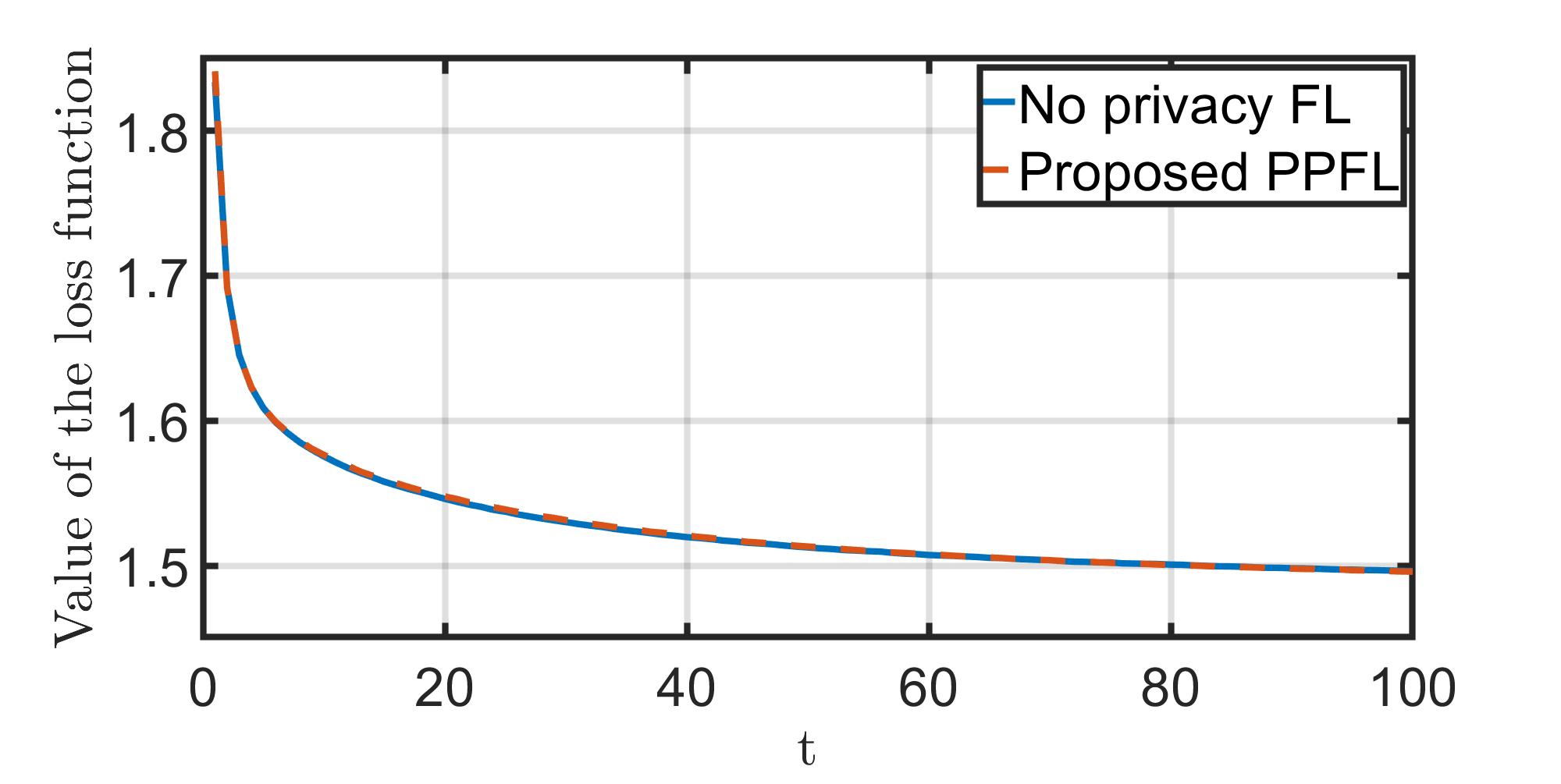}
  \caption{The comparison of the loss function amount of FL network with and without the proposed privacy mechanism.}\label{fig:loss}
\end{figure}
%%%%%%%%%%%%%%%%%
\indent We design matrix $M$ randomly as the decryption key of SIFL with dimension $n \times (n+1)$, where $n=199,210$ is the total number of model parameters in the MLP network. Next, we calculate its inverse and null space, $G$ and $N$. Vector $R^t$ is randomly chosen in every global iteration.\\
%?????Therefore at each communication round, the clients’ models are trained based on encrypted global model individually before being averaged into the aggregated model.\\
\indent The comparison of training accuracy and loss results of SIFL framework and the standard FL without privacy are shown in Figure \ref{fig:accuracy} and \ref{fig:loss}, respectively. As can be seen, the accuracy and the evolution of the loss function with the SIFL setting are almost the same as the accuracy and loss with no privacy setting, which shows that SIFL can integrate a cryptographic method in the FL system without sacrificing model accuracy and convergence rate. Therefore, there is no need to hold a trade-off between privacy and the performance of FL.\\
\indent In the solution section, Section \rom{3}, we show that in the SIFL framework, the encrypted local model of each client is an affine function of the original local model. This is shown in Figure \ref{fig:Normpflocal model} based on the second norm of local model parameters as $||\tilde{w}_i^{t+1}||=||G w_i^{t+1} + NR^t||$ for the first client and in all iterations. Also, in Figure \ref{fig:errorlocalmodel}, the norm of the relative error between $\tilde{w}_i^{t+1}$ and $(G w_i^{t+1} + NR^t)$ is depicted. The error between updated weights and the expected updated weights is caused by the huge dimension of the parameter vector in this case and the matrix inversion in the calculation of the encryption key, which leads to some calculation errors.\\
\indent Finally, in Figures \ref{fig:encryptdecrypttime} and \ref{fig:timecost}, we investigate the effect of encryption and decryption operations in  SIFL on the training time of FL. As can be seen, the increased training time compared to the training time of the original FL is negligible.
%%%%%%%%%%%%%%%%%%%%%%%%%%%%%%
%% norm of local model

%% norm of relative error

%% encrypt decrypt time

%% time cost
%%%%%%
%In this section, we introduce the privacy-preserving FL framework that we propose. This framework is built on matrix encryption and system immersion tools from control theory. We refer to it as System Immersion Federated Learning (SIFL). SIFL provides the same accuracy and convergence rate as the standard FL, reveals no information about clients' data, and is computationally efficient. It provides high level of privacy without degrading the performance of FL.
%%%%%%%%%%%%%%%
\section{CONCLUSIONS}
In this paper, we proposed a System Immersion Federated Learning, SIFL, as a privacy-preserving FL framework built on the synergy of matrix encryption and system immersion tools from control theory to provide unconditional secrecy for the clients' data in federated learning. As a privacy mechanism, we developed an immersion for the learning algorithm (SGD), and we designed the dynamics of a target system so that trajectories of the original SGD are immersed in its trajectories, and it learns on encrypted data based on the random matrix encryption. Matrix encryption was reformulated at the server as a random change of coordinates that maps original parameters to a higher-dimensional parameter space and enforces that the target SGD converges to an encrypted version of the original SGD optimal solution.\\
\indent SIFL provides the same accuracy and convergence rate as the standard FL, reveals no information about clients' data, and is computationally efficient. It provides a high level of privacy without degrading the performance of FL. The simulation results of SIFL are presented to illustrate the performance of our tool. These results demonstrate that SIFL provides the same accuracy and convergence rate as the standard FL with a negligible computation cost.
%%%%%%
%\begin{appendices}
%\section{Proof of Proposition 1}
%Proof of $\tilde{w}_i^{t+1}=G w_i^{t+1} + NR^t w^t$:\\
%Server sends $(\tilde{w}^t=A^t w^t=G w^{t}+N R^{t} w^{t})$ to clients.\\
%In each round of FL, epoch iterations of SGD are run on each active client. The output of the first iteration of SGD can be calculated as follows:
%\begin{align}
%\tilde{z}_i^{t+1}&=\tilde{w}^{t}-\eta G \nabla l\left(M \tilde{w}^{t},D_i\right)\nonumber\\
%&=G w^{t}+N R^{t} w^{t}-\eta G \nabla l\left(w^{t},D_i\right)\\
%&=G\left(w^{t}-\eta \nabla l\left(w^{t}\right)\right)+N R^{t} w^{t}=G z_i^{t+1}+N R^{t} w^{t},
%\end{align}
%where $\tilde{z}_i^{t+j}$ and ${z}_i^{t+j}$ are the outputs of SGD after $j$-th iterations with and without privacy distortion, respectively. Then, the output of SGD after second iteration is:
%\begin{align}
%\tilde{z}_i^{t+2}&=\tilde{z}_i^{t+1}-\eta G \nabla l\left(M \tilde{z}_i^{t+1},D_i\right)\\
%&=G {z}_i^{t+1} + NR^{t}{w}^{t} - \eta G{\nabla l}\left(z_i^{t+1},D_i\right)\\
%&=G\left({z}_i^{t+1}-\eta G \nabla l\left({z}_i^{t+1},D_i\right)\right)+NR^{t} w^{t}\\
%&=G z_i^{t+2}+NR^{t} w^{t}.
%\end{align}
%Therefore, by induction, the output of SGD after $epochs=E$ iterations is as follows:
%\begin{equation}
%    {\tilde{z}_i^{t+E}=G \tilde{z}_i^{t+E}+NR^{t} %w^{t}} = \tilde{w}^{t+1}.
%\end{equation}
%Then, we can conclude that:
%\begin{equation}
%    \tilde{w}_i^{t+1}=G w_i^{t+1} + NR^t w^t.
%\end{equation}
%\hfill $\blacksquare$
%\end{appendices}
%%%%%%%%%%%%%5555
%%%%%%%
\bibliographystyle{IEEEtran}
\bibliography{conference_101719}
%%%%%%%%%%%%%%%%%%%%%%%%%%%%%%5
\end{document}